%% file: main.tex
\documentclass[11t]{article}
\usepackage{fullpage}
\usepackage[utf8]{inputenc}
\usepackage[round]{natbib}  
\usepackage{authblk}
\usepackage[table]{xcolor}
\usepackage{amsmath}
\usepackage{graphicx}

\newenvironment{summaryquote}
  {
   \quote 
   \singlespacing 
   \parindent 1em 
  }
  { 
   \endquote 
  }

\emergencystretch=\maxdimen
\usepackage{longtable}
\usepackage{rotating}
\usepackage{setspace}
\usepackage{multirow}
\usepackage{subcaption} 
\usepackage{times}
\usepackage{latexsym}
\usepackage{multirow}
\usepackage{todonotes}
\usepackage{xcolor}
\usepackage{soul}
\usepackage{hyperref}
\usepackage{bibliography_sty}

\author[1]{Elliott Ash}
\affil[1]{\small ETH Zurich}

\author[2]{Aniket Kesari}
\affil[2]{\small Fordham University}

\author[3]{Suresh Naidu}
\affil[3]{\small Columbia University}

\author[4]{Lena Song}
\affil[4]{\small University of Illinois at Urbana-Champaign}

\author[1]{Dominik Stammbach}

\title{{\huge {\textbf{Translating Legalese}} \\  Enhancing Public Understanding of Court Opinions with Legal Summarizers}}

\date{\today}

\begin{document}

\maketitle

\begin{abstract}
    \singlespacing \noindent Judicial opinions are written to be persuasive and could build public trust in court decisions, yet they can be difficult for non-experts to understand. We present a pipeline for using an AI assistant to generate simplified summaries of judicial opinions. Compared to existing expert-written summaries, these AI-generated simple summaries are more accessible to the public and more easily understood by non-experts. We show in a survey experiment that the AI summaries help respondents understand the key features of a ruling, and have higher perceived quality, especially for respondents with less formal education. 
     
\end{abstract}

\clearpage 
\setcounter{page}{1}

\section{Introduction}\label{intro}



Judges are important policymakers but are less accountable to the public than legislators.\footnote{There is a rich literature on the role of courts as policymakers. Although judges often disclaim their role as policymakers (e.g., Chief Justice John Roberts famously saying that judges simply ``call balls and strikes''), scholars have long argued that these characterizations are inaccurate. Courts act as political institutions and are important in not only interpreting, but actually shaping public policy \cite{feeley2000judicial, epstein2001supreme}.} One way that judges strengthen the legitimacy of their policy choices given low accountability is by providing written justifications based on shared principles, which are then published as judicial opinions. John Rawls argued that ``[The U.S. Supreme Court's] role is not merely defensive but to give due and continuing effect to public reason by serving as its institutional exemplar'' \cite{Rawls1993-RAWPL}. Presumably, this legitimizing function is best served when the general population can understand the written justifications. In practice, however, judicial opinions tend to be extremely long and written in complicated technical language that is inaccessible except to trained lawyers. This use of ``legalese'' makes it impossible for non-experts to read and comprehend such opinions. The resulting gap in accessibility frustrates the legitimizing function of published judicial opinions and potentially contributes to gaps in public trust in the courts.

Recognizing the importance of explaining high-profile decisions to the public, there are several efforts to summarize court opinions for public consumption. Resources such as SCOTUSblog, Oyez, and Wikipedia all provide summaries for a few cases. However, only highly trained legal experts can write such summaries, which makes them costly to produce. And even for trained lawyers, writing such a summary is an expensive and time-consuming process. Consequently, only the most high-profile cases have been summarized for the public. To make matters more complicated, even when such summaries are available, they are often still too sophisticated for most readers. Moreover, the summaries often lack additional important information, such as the procedural context or the opinions given by dissenting judges.

Recent advances in natural language processing and machine learning make it possible to automatically generate such summaries at scale. Large language models (LLMs) such as GPT \cite{openai2023gpt4} are capable of automatically conducting many summarization tasks across many domains. \citet{pu2023summarization} show that such generated summaries often are preferred to human equivalents. LLMs have also been scrutinized in a number of legal contexts. For example, \citet{Katz2023-bw} show how GPT-4 successfully passed the bar exam.

We explore these models' potential to effectively summarize court opinions. We present a pipeline for automatically generating high-quality summaries of court opinions. The pipeline uses GPT-4 to extract and summarize the facts of the case and the principal arguments used in the majority opinion. We further illustrate that the style of the summaries can be targeted for specific contexts and audiences, such as for social media or news articles. We combine this style transfer with a simplification step, which in turn makes the summaries more accessible to a non-technical audience.

We validate the accessibility and quality of our summaries with a survey experiment. For a set of U.S. Supreme Court opinions, we ask participants to read either an expert-written summary or an AI-generated simple summary written at the seventh-grade level. We show that with access to the simple summary, respondents can better answer basic factual questions about the case. Respondents also report higher qualitative ratings of the simple summaries (compared to the expert-written summary) and are more likely to share the summary with a friend or relative. These preferences for the AI summaries are even larger for respondents with relatively low levels of formal education.

Our procedure combines evaluation of machine-generated summaries with established practices from experimental social sciences. A trained legal professional who is sufficiently familiar with all summarized cases has read all generated summaries and is satisfied with their quality, in particular faithfulness to the original opinion's arguments. We then use a survey experiment to investigate whether the summaries help make the law more accessible to non-experts. This novel protocol emphasizes the evaluation of the impact of such summaries on a target audience, instead of assessing quality using metrics such as word overlap, readability, or fluency \citep{pu2023summarization} that measure text quality more generally.

Taking a use-case-dependent and user-centric evaluation approach highlights another limitation of \textit{quality-based} evaluation of summaries: There exists an inherent precision-accessibility trade-off that has not been explored widely in the natural language processing literature. Our simplification step trades off lower precision in the summaries for more accessibility. The survey results confirm that the accessible summaries are preferred by our survey participants, while also improving understanding of the opinions.

This paper proceeds as follows. Section \ref{related_work} describes related work. Section \ref{legal_summarizer} describes the development of an NLP pipeline for effective case summarization. Section \ref{survey} describes the survey experiment. Section \ref{results} describes the main results and Section \ref{discussion} provides a concluding discussion.

\section{Background: Simple Summaries}\label{related_work}


Automated summarization can be either achieved via extractive or abstractive methods. In extractive summarization, we are interested in extracting keywords, phrases, sentences, or paragraphs from a lengthy piece of text. Such approaches have already been explored to summarize Supreme Court opinions \citep{bauer2023legal}. In abstractive summarization, we generate a paraphrased summary from scratch using the source document as a reference, similar to machine-translation approaches where the translated text is generated from scratch given the source.
Extractive summaries can be too verbose, and the excerpts might lack coherence on their own. Abstractive summaries are usually more coherent, but they are prone to hallucination -- altering or inventing facts or language -- which can be especially troublesome in a technical and high-stakes field like law.

The question of what constitutes a good summary is not settled  \citep{hahn_2000}, apart from that the summary should be shorter and more concise than the source \citep{long_document_summarization}. But in the case of technical documents, short summaries may require additional explanation of key terms not present in the source document. Moreover, different target audiences might evaluate the same summary differently, depending on their domain knowledge, use case, and preferences. \citet{krishna-etal-2023-longeval} discuss best practices to evaluate faithfulness, but good summaries have many more dimensions, most of which are context- and user-specific.

Notwithstanding these evaluation issues, the technology of summarization has developed rapidly. \citet{stiennon_2020} show that the GPT-3 model \citep{brown_et_al_gpt3} can effectively and accurately perform zero-shot summarization when aligned with human feedback. They argue it is hard to formalize what a good summary consists of, but effective summaries are recognizable (\textit{I know it when I see it}). For example, \citet{stammbach_ash} combine dimensions of summarization, style transfer, and explanations into a single summary in the context of explainable fact-checking, using an abstractive summarization approach and GPT-3.  Lastly, \citet{goyal2023news} and \citet{pu2023summarization} show that zero-shot summaries produced by large language models are often preferred by humans compared to state-of-the-art dedicated summarization systems \citep[e.g.,][]{liu-etal-2022-brio}.

Our pipeline incorporates elements from various strands of existing research. These include summarizing U.S. legal documents \citep[e.g.,][]{billsum, bauer2023legal}, style transfer approaches \citep[e.g.,][]{krishna-etal-2020-reformulating, feng2023sentence}, and explanation-producing systems in other contexts \citep{kumar-talukdar-2020-nile, atanasova-etal-2020-generating-fact, stammbach_ash}. We explored the use of independent modules and pipelining them together, but found that such local approaches do not work well for summarizing judicial opinions. First, these models show a limited understanding of language in general and, more importantly, do not translate well to the legal domain (as domain shifts are a major obstacle in natural language processing). Moreover, such supervised methods require the existence of high-quality and domain-specific training resources, which do not exist for our purposes and would be infeasible to create.

This points toward leveraging large language models such as GPT-4 \cite{openai2023gpt4} for summarization and style transfer. As GPT has recently gained more popularity, increasing attention is being paid to ``prompt engineering.'' Writing a suitable prompt is crucial for extracting the desired behavior from large language models \citep[see,e.g.][]{schick2021its, prompt_programming, kojima2023large} In our work, we find that these observations hold in the field of law as well.

\section{Simple Summaries of Legal Opinions}\label{legal_summarizer}

\begin{figure}
    \centering
    \caption{System Overview: Producing Simple Summaries}
    \label{fig:flowchart-summarization}
    \includegraphics[width=0.5\textwidth]{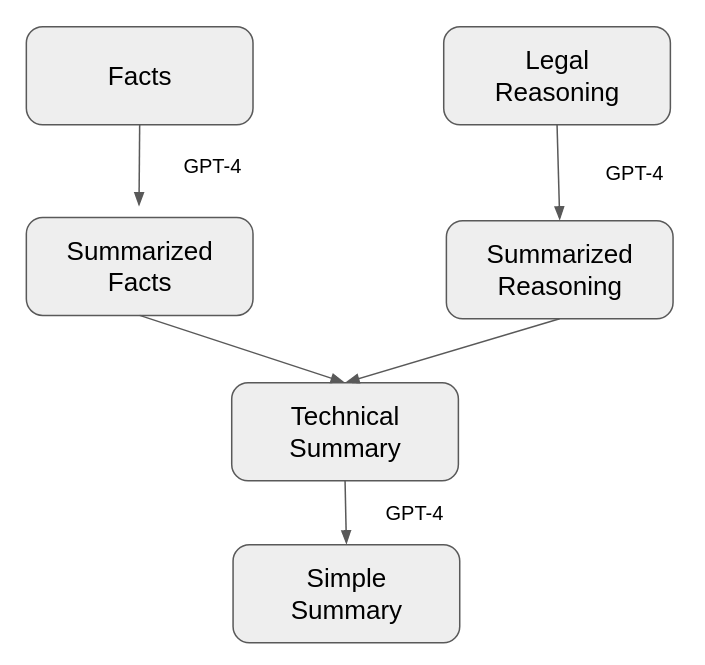} 
    
\end{figure}

At a high level, our legal summarizer pipeline takes the text of a Supreme Court opinion and returns a short summary in a given style. We provide an overview in Figure \ref{fig:flowchart-summarization}. We provide methods for summarizing opinions as a Twitter thread, a YouTube comment, or a short essay at a 7th-grade reading level. Other styles can be relatively easily implemented by adjusting the LLM prompts. For the summarization tasks, we use OpenAI's GPT-4 model 4096 max input and output tokens, 0 temperature). Our pipeline generates a series of longer-form summaries: one summary for the facts of a case, and one summary for the legal reasoning. We then concatenate all these summaries together. This serves as input to produce a final, accessible, style-transferred summary.\footnote{We also experimented with doing summarization and style-transfer in one step. Because of context window constraints, this required an additional summarization step regardless. We therefore left the style-transfer to the last step. With a larger context window, it may be possible to combine the summarization and style transfer steps.} This procedure resembles recursive summarization approaches described in  \cite{wu2021recursively}.

\subsection{Data}

For our application, we use the text from U.S. Supreme Court opinions. We selected 15 cases from the last fifteen years based on notability and the topics covered. We retrieve the majority opinion syllabi and the facts of the case from Oyez. We picked cases covering affirmative action\footnote{\textit{Schuette v. Coalition to Defend Affirmative Action} (2014), \textit{Fisher v. University of Texas} (2016), \textit{Students for Fair Admissions v. North Carolina/Harvard} (2023).}, abortion\footnote{\textit{Gonzales v. Carhart} (2007), \textit{Whole Woman's Health v. Hellerstedt} (2016), \textit{Dobbs v. Jackson} (2022).}, search and seizure\footnote{\textit{U.S. v. Jones} (2012), \textit{Riley v. California} (2014), \textit{Carpenter v. U.S.} (2018).}, labor\footnote{\textit{Janus v. AFSCME} (2018), \textit{Harris v Quinn} (2014), \textit{Glacier Northwest, Inc. v. International Brotherhood of Teamsters} (2023).}, and LGBT rights\footnote{\textit{U.S. v Windsor} (2013), \textit{Obergefell v. Hodges} (2015), \textit{Bostock v. Clayton County} (2020).}. Each of these cases is among the most high-profile decisions within their topics and generally does not implicate more technical areas of law.    

\subsection{Summarization}

As outlined in Figure \ref{fig:flowchart-summarization}, the first step is to summarize all the facts of a case. We use GPT-4 and prompt it with the following instruction:
\begin{quote}
        \singlespacing Take this summary of the facts of a case for a U.S. Supreme Court case and simplify it into 1-2 sentences.
\end{quote}
followed by the text of the facts for a given case.

We save the output of this step, which will be needed later. We then proceed by summarizing the court-provided syllabus of the Supreme Court majority opinion. We use the following prompt, which was developed in an iterative manner until adequate quality was achieved.\footnote{We began with a simple prompt ``Highlight the key arguments from the following text from a U.S. Supreme Court opinion syllabus.'' We then added the requirement to provide a high-level summary as it was clear the syllabi did not always contain this information (though opinions usually do) and would be necessary for respondents to understand what the case is about. We found that these summaries would often produce text like ``Chief Justice Roberts said...'' or ``Justice Sonia Sotomayor believes...'' and added the requirements to write in third person and anonymously so that respondents would focus on the content of the summary and not the identity of the Justice. We found that these summaries could end up introducing counterarguments or commentary about the decision, so we added the requirements to write persuasively and like a Supreme Court justice in order to present the opinion faithfully. Finally, we added the requirement to define jargon to ease comprehension by lay readers as these terms are not always defined in the syllabi or opinions.}
\begin{quote}
            \singlespacing Highlight the key arguments from the following text from a U.S. Supreme Court opinion syllabus in 2000 words or fewer from the perspective of the majority. Make sure the beginning gives a high-level summary of what the case is about (e.g. basic facts of the case, area of law, etc.).  Write in third person (for example, `the law requires...'), while also making sure to anonymize the identity of the author of the opinion. Write this summary in a way to persuade a reader to agree with the logic and conclusion. Make sure to maintain a serious tone appropriate for the Supreme Court.  For any legal jargon (such as `penumbras,' `incorporation,' `Miranda rights,' or `strict scrutiny'), add a * next to the word or phrase, then at the bottom of the summary, define the term.
\end{quote}

We experimented with summarizing both the court-provided syllabus and the full text of the majority opinion. Because GPT-4 limits documents to roughly 6000 words (8192 tokens) including both the input and output, long inputs have to be chunked into smaller pieces before summarization. Because the full majority opinions are usually longer than 6000 words, we use the syllabus as the default. Note that the prompt also works well with chunks of the full opinions as input. When chunking, a summary is generated for each piece, and then each summary is appended to the same list along with the facts summary.

Outputs can be sensitive to the system prompt. To illustrate this point, consider the differences in outputs when summarizing \textit{Dobbs v. Jackson Women's Health Organization} with different prompts. Take, for example, this shorter prompt:
\begin{quote}
    \singlespacing Highlight the key arguments from the following text from a U.S. Supreme Court opinion syllabus.
\end{quote}
This prompt produces a 2200-word summary. Looking at just the start of the summary (emphasis added):
\begin{summaryquote}
    \singlespacing The key arguments in this text are:

1. Mississippi's Gestational Age Act, which prohibits abortions after 15 weeks of gestation except in cases of medical emergency or severe fetal abnormality, was challenged by Jackson Women's Health Organization and one of its doctors. They argued that the Act violated the constitutional right to abortion as established by the Supreme Court's precedents in Roe v. Wade and Planned Parenthood of Southeastern Pa. v. Casey.

2. The District Court ruled in favor of the respondents, stating that Mississippi's 15-week restriction on abortion violates the Supreme Court's cases that forbid states from banning abortion pre-viability. This decision was affirmed by the Fifth Circuit.

3. The petitioners defended the Act before the Supreme Court, arguing that Roe and Casey were wrongly decided and that the Act is constitutional because it satisfies rational-basis review.

4. The Supreme Court held that the Constitution does not confer a right to abortion, overruled Roe and Casey, and returned the authority to regulate abortion to the people and their elected representatives. The critical question, according to the Court, is whether the Constitution, properly understood, confers a right to obtain an abortion.
The key arguments in this text are:
\end{summaryquote}
This summary can be confusing for lay readers in several ways. First, it gives little in the way of factual background of the case. In some cases, we found this kind of prompt would not mention the facts at all, and instead launch into the summary of the opinion. Second, the summary does not always adequately define legal jargon that might be inaccessible to lay readers. Terms like ``petitioners'' and ``rational-basis review'' may not be understood by lay audiences. While readers may be able to broadly understand the core arguments of conservative and liberal justices on high-profile issues such as abortion, undefined jargon may inhibit lay understanding of technical legal arguments. 

Each addition to the base prompt corrects some default behavior that might inhibit public understanding of the court opinion. We make sure the summary starts with the facts of the case so that readers know what the controversy is about, even for less high-profile cases. Writing in third person, anonymously, and persuasively ensures that the summary reflects the court's opinion as closely as possible while focusing the reader's attention on the legal arguments, rather than personal factors. It is important to note however, that a large language model can easily be forced to output content that aligns with some particular viewpoint. These so-called ``spinning attacks'' could be used to summarize opinions in such a way that agree with the prompter's priors \cite{Bagdasaryan2022}.  

While jargon may be unavoidable, defining it helps readers follow along with the core legal arguments. Notably, GPT-4 is able to define jargon that it is not specifically instructed to. In our prompt, we only mention `penumbras,' `incorporation,' `Miranda rights,' or `strict scrutiny' as examples, yet the model analogizes to define other similar legal jargon. For instance, adding this language:

\begin{quote}
    For any legal jargon (such as `penumbras,’ `incorporation,’ `Miranda rights,’ or `strict scrutiny’), add a * next to the word or phrase, then at the bottom of the summary, define the term.
\end{quote}

to the prompt produces this text:
\begin{summaryquote}
    \singlespacing The case in question involves the Mississippi Gestational Age Act, which prohibits abortions after 15 weeks of gestational age, except in cases of medical emergencies or severe fetal abnormalities. The Act was challenged by Jackson Women's Health Organization and one of its doctors, who argued that it violated the constitutional right to abortion as established by the Supreme Court precedents Roe v. Wade and Planned Parenthood of Southeastern Pa. v. Casey. The District Court ruled in favor of the respondents, and the Fifth Circuit affirmed this decision, both arguing that the Act violated the prohibition against pre-viability abortion bans. The petitioners, however, defended the Act by arguing that Roe and Casey were wrongly decided and that the Act is constitutional because it satisfies rational-basis review*.

    The key argument of the Supreme Court's decision is that the Constitution does not confer a right to abortion. Therefore, the precedents set by Roe and Casey are overruled, and the authority to regulate abortion is returned to the people and their elected representatives. The Court argues that the critical question is whether the Constitution, properly understood, confers a right to obtain an abortion. The controlling opinion in Casey did not address this question and reaffirmed Roe solely on the basis of stare decisis*. However, a proper application of stare decisis requires an assessment of the validity of the precedents in question.

    *Rational-basis review: A standard of judicial review that examines whether a legislature had a reasonable and not an arbitrary basis for enacting a particular statute.\newline
    *Stare decisis: The legal principle of determining points in litigation according to precedent.
\end{summaryquote}

Despite not being explicitly instructed to do so, the model defines popular legal terms such as rational-basis review and stare decisis, It also defines terms specific to the case, such as in this passage that references the fetal viability standards established in \textit{Roe} and \textit{Planned Parenthood}:

\begin{summaryquote}
    \singlespacing The author criticizes the Casey decision for abandoning the privacy right basis of Roe v. Wade and grounding the abortion right entirely on the Fourteenth Amendment's Due Process Clause. The Casey decision also rejected Roe's trimester scheme* and replaced it with the ``undue burden'' test, which the author argues lacks firm grounding in constitutional text, history, or precedent.

    The author further argues that the "undue burden" test is unworkable, as evidenced by its inconsistent application in lower courts and the numerous Circuit conflicts it has generated. The author suggests that continued adherence to this test would undermine, rather than advance, the rule of law.

    *Trimester scheme: A framework established by Roe v. Wade that divided pregnancy into three trimesters and set different standards for state regulation of abortion in each trimester.
\end{summaryquote}

These supporting points are very helpful in understanding the legal and practical significance of these documents.

\subsection{Style Transfer}

Even with effective summarization, there is still a remaining question of how to convey these summaries to non-legal audiences. The summaries produced by our summarization step still tend to be long (often upward of 3000 words) and written at a fairly sophisticated reading level. To solve this problem, we provide a method for taking these long summaries and summarizing them again through a style transfer. Style transfer is commonly used in other domains such as computer vision (e.g. changing a photograph to be in the style of van Gogh). In this case, we use it to transfer from one text style (Supreme Court opinions) to another (e.g. a YouTube comment). We provide methods for transferring the initial summary into Twitter threads, YouTube comments, or 7th-grade-level essays. 

To accomplish this step, we take the output from the summarization step and then put it through the GPT-4 model with the following prompt (swapping the appropriate output style):
\begin{quote}
            \singlespacing Take this summary of a Supreme Court opinion and summarize it in 10 short paragraphs or fewer at a \textbf{7th-grade reading level}. Number each paragraph at the start like 1), 2), 3), etc. Make sure the first paragraph gives a high-level summary of what the case is about (e.g. basic facts of the case, area of law, etc.).  Write in third person (for example, `the law requires...'). Write this summary in a way to persuade a reader to agree with the logic and conclusion. For any legal jargon (such as `penumbras,' `incorporation,' `Miranda rights,' or `strict scrutiny'), add a * next to the word or phrase, then at the bottom of the thread, define the term.
\end{quote}
We still include the instructions from the previous step to define legal jargon and write persuasively, while adding instructions to format the output with numbered paragraphs for readability. 

We show the start of a style-transfered summary to 7th-grade reading level below and show all full summaries in Appendix \ref{app:style-transfer-summaries}.

\begin{summaryquote}
    \singlespacing 1) In 2018, a law in Mississippi, known as the "Gestational Age Act," was challenged. This law banned most abortions after 15 weeks. The challenge came from the Jackson Women's Health Organization, the only licensed abortion clinic in the state. Lower courts blocked the law, saying there wasn't enough evidence that a fetus could survive outside the womb at 15 weeks. They also said that prior Supreme Court rulings didn't allow bans on abortions before a fetus could survive on its own.

2) The Supreme Court was asked to look at this case. They had to consider whether the Constitution, the highest law in the land, gives women a right to get an abortion. They decided that the Constitution does not. They disagreed with earlier decisions in the cases of Roe v. Wade and Planned Parenthood of Southeastern Pa. v. Casey, which had said that the Constitution did give women this right.

3) The Supreme Court also looked at whether the right to get an abortion is part of the "liberty" that the Constitution protects. The Constitution doesn't directly say that women have a right to get an abortion. But in the past, the Supreme Court had said that this right was part of a broader right to privacy. The new decision said that the Constitution doesn't protect this right.

4) ...
\end{summaryquote}

Each style-transferred summary maintains the behavior of defining legal jargon and maintaining the majority's viewpoint, while also being more concise. As compared to the 2000-4000 word summaries generated by the first step, the style transfer step distills the \textit{Dobbs} summary significantly -- to ca. 600 words for the 7th-grade reading level style, for example.

\section{Survey Experiment}\label{survey}

To evaluate our summaries, we conducted a survey experiment assessing reader comprehension and assessment of our summaries relative to expert-written summaries. In July 2023, we surveyed 120 survey respondents recruited from Prolific, stratified across educational level: Half of the survey participants have a college degree or higher, the other half do not have a college degree. We collect standard demographic information about race, sex, age, and political party, as well as standard attention checks. 

Each survey respondent was asked to read five different passages about Supreme Court decisions across five topic areas (affirmative action, abortion, labor, search and seizure, or LGBT rights). We randomly drew one case from each category. Each category contains three cases that can potentially be drawn and shown to a participant. We show all 15 cases considered in our experiment in Appendix Table \ref{tab:selected-cases}.

For each case, respondents were randomly assigned to either the text of a Justia summary, a concise summary drafted by legal professionals\footnote{\url{https://en.wikipedia.org/wiki/Justia}} (the control group), or to our GPT-4 generated 7th-grade-level summary of the same case (treatment group). We asked survey participants to read the text carefully. Participants were not given any incentives beyond payment for completing the survey. 

In the first set of questions, we asked two standardized recall questions about the case: a multiple-choice question about the area of law for a case (affirmative action, abortion, labor, search and seizure, or LGBT rights) and a binary-choice question about the general direction of the decision, based on the legal area. That is, depending on the case, we ask if the judges favor/oppose affirmative action, abortion rights, collective bargaining, warrantless searches, or LGBT rights. If our summaries support a better understanding of the decision, we expect a higher rate of correct answers when the respondents have access to them.

Second, we ask more subjective questions about the quality or usefulness of the simple summaries. We ask them to assess the level of detail (too much, too little, just right) and the clarity of the main points of the text. Finally, we ask the respondents whether they would forward the text to an interested friend or relative to help them understand the case.%

\section{Results}\label{results}

In this section, we summarize our results. We show that our GPT-4 generated summaries and baselines differ in reading difficulty, and our summaries are more accessible to non-experts than short expert-written summaries. 

\subsection{Descriptive Statistics}

\begin{figure}
\caption{Readability Statistics Across Summary Types}\label{fig:quant-metrics}

\centering
\begin{minipage}{.35\textwidth}
\centering 
\caption*{A. Different Summary Stages}
\includegraphics[width=\textwidth]{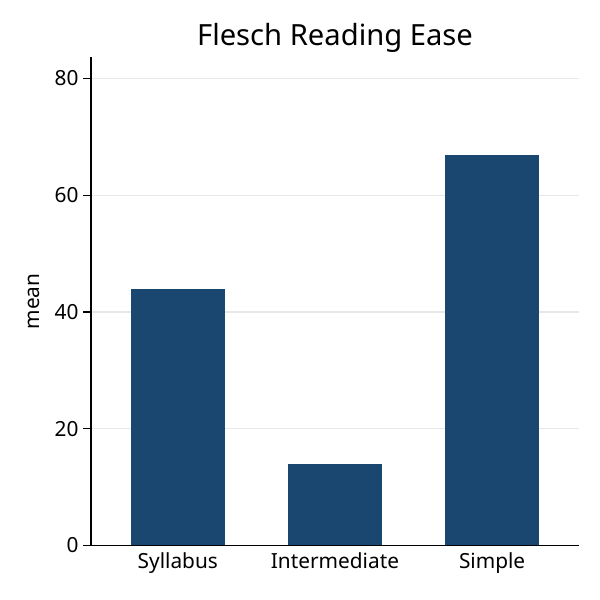}
\label{fig:flesch-summary}
\end{minipage}%
\linebreak
\begin{minipage}{.35\textwidth}
\centering 
\caption*{B. GPT-4 vs. Justia Summaries}
\includegraphics[width=\textwidth]{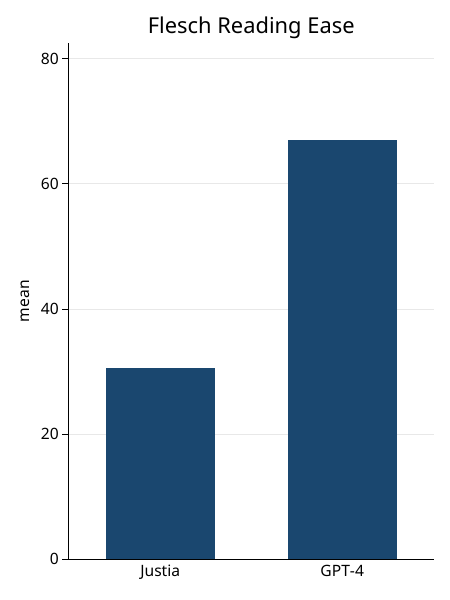}
\label{fig:flesch-justia}
\end{minipage}

\flushleft Note: This figure shows the reading difficulties for the original syllabi, intermediate summaries and simplified summaries in Panel A, and the simplified GPT-4 summaries and Justia summaries in Panel B.
\end{figure}

In Figure \ref{fig:quant-metrics}, we show reading difficulties of different text, using Flesch Reading Ease scores \cite{flesch1948new}. Flesch scores were developed to evaluate reading ease, ranging from a 5th-grade reading level up through professional, and have been used to evaluate legal documents. For example, states oftentimes have laws requiring that certain documents like insurance policies conform to a certain grade-reading level and Flesch scores have been used to assess compliance. Flesch scores are calculated via the formula displayed in Equation \ref{eq:reading_ease}.

\begin{equation} \label{eq:reading_ease}
   y = 206.185 - 1.015 * \frac{\text{total words}}{\text{total sentences}} - 84.6 * \frac{\text{total syllables}}{\text{total words}}
\end{equation}

A text containing relatively long sentences and words with lots of syllables has a low readability score. Conversely, text with relatively short sentences and words with few syllables is easy to read and has a high readability score. In Figure \ref{fig:quant-metrics}, we show these readability scores for the Supreme Court syllabi of our cases, the intermediate compact summary, and the style-transferred simple version of the summary. We also show readability scores for Justia summaries, which act as the control in our survey experiment.

We find that Supreme Court syllabi have a readability score of around 40, which corresponds to \textit{hard to read} or college-level text. Summarizing Court opinions results in intermediate summaries with a Flesch score of around 15, which corresponds to \textit{very difficult to read} text, best understood by university graduates. While GPT-4 condenses the Supreme Court opinions into short texts, these texts consist of a highly condensed version of the source (as most summaries), which is hard to read. In contrast, applying our style transfer to these intermediate summaries results in readability scores of 65, which can be interpreted as \textit{plain English} which is easily understood by 13-15-year-old students. Finally, the Justia summaries written by professionals result in a Flesch score of 30, which again is \textit{very difficult to read} text and best understood by university graduates. 

These scores give us a rough indication that the surface form of the generated easy-to-read summaries should be easily understood by lay people, and thus might make Supreme Court opinions more accessible. However, such scores are calculated based on simple statistics and do not reflect any information about the content of the generated summaries. Hence, we run a survey experiment presenting the information provided in the simple summaries compared to a strong baseline.

\subsection{Evaluation Results}

We evaluate the effect of reading our AI summaries against Justia summaries using Equation \ref{eq:survey_results}:
\begin{equation}\label{eq:survey_results}
    Y_{jk} = \alpha_{k} + \beta\text{Treated}_{jk} + \epsilon_{jk}
\end{equation}
where $\alpha_{k}$ are case fixed effects, $\text{Treated}_{jk}$ equals 1 if participant $j$ was assigned to read an AI-summary for case $k$ and 0 if the participant was assigned to read the control summary. Standard errors $\epsilon_{jk}$ are clustered by respondent $j$. This allows us to identify how GPT-generated summaries compare relative to the control Justia summaries, which is a strong baseline and already more accessible than the original judicial opinions.

\begin{figure}
\caption{Main Results: Effects on Comprehension and Perceived Quality.} \label{fig:understanding_summaries_impact}
\centering

\begin{minipage}{.33\textwidth}
\caption*{A. Reading Comprehension}
\includegraphics[width=\textwidth]{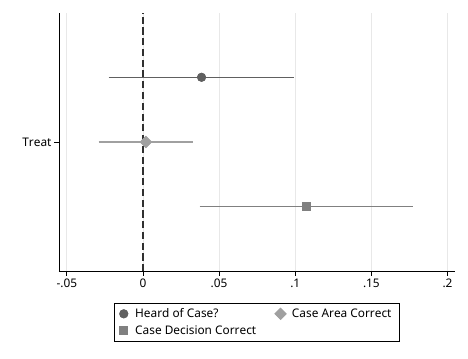}
\end{minipage}%
\linebreak
\begin{minipage}{.33\textwidth}
\caption*{B. Perceived Quality}
\includegraphics[width=\textwidth]{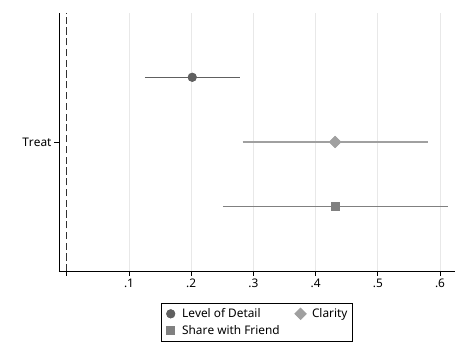}
\end{minipage}%
\flushleft {\footnotesize Notes. This figure presents the treatment effect of being exposed to our AI summaries, compared to control (Justia summaries). Error bars reflect 95 percent confidence intervals. We find that survey participants exposed to our summaries are more likely to guess the correct case decision and believe GPT-4 summaries more often have the right amount of detail and are clearer compared to the Justia summaries. Participants are also more likely to share GPT-4 summaries with friends compared to Justia summaries. } 
\end{figure}

\begin{figure}
\centering

\caption{Effects By Education Level.} \label{fig:understanding_summaries_impact}
\begin{minipage}{.33\textwidth}
\includegraphics[width=\textwidth]{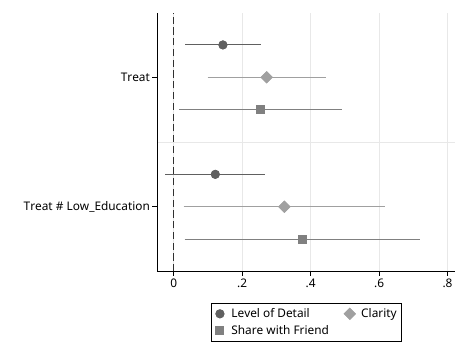}
\end{minipage}%

\flushleft {\footnotesize Notes. This figure presents treatment effects on perceived quality by education level. Error bars reflect 95 percent confidence intervals. Quality effects are larger for respondents with lower education.} 
\end{figure}

Figure \ref{fig:understanding_summaries_impact} presents our main results.\footnote{The results in these two panels are summarized in Appendix Table \ref{tab:main}.} In the first coefficient plot (Panel A), we ask background and reading comprehension questions, including whether participants have already heard of a case, whether they identified the correct area of law, and whether they can identify the correct outcome of the decision (such as whether a labor case is in favor of / opposition to unions relative to employers).  We find a null effect on whether people have heard about a given case, which should not be influenced by the summary a participant is exposed to. There is also no effect on whether participants correctly identified the area of law, as this compensation question is simple and answered accurately 96\% of the time. In contrast, the AI summary significantly improves participants' answers on the case decision question. On average, participants correctly identified the case decision 74\% of the time. Participants who were exposed to the GPT-generated summaries correctly labeled the overall direction of Supreme Court opinions around 80\% of the time, an 11\% increase relative to control participants exposed to Justia summaries. That is, the AI summaries make decisions significantly more accessible.

In Figure \ref{fig:understanding_summaries_impact} Panel B, we show perceived quality of the summaries by survey participants. We find significant effects in all questions asked. Survey respondents more often assess the GPT-generated summaries to have the right level of detail (not too much or too little) and the right amount of clarity. Respondents also would share GPT-generated summaries more often with friends or families if they think a case is important.  Across three different measurements, we see that the AI summaries are systematically preferred to human-generated expert summaries.

In the last coefficient plot in Figure \ref{fig:understanding_summaries_impact}, we show these subjective quality effects by education level. We interact the treatment indicator with an indicator for non-college education (no college degree or lower). We find significant interaction effects. While the AI summaries are assessed positively by all respondents, it is especially among participants who have not attended college where the GPT-generated summaries are interpreted as more useful than Justia summaries. That is, they assess the AI summaries to be clearer and are more likely to share these AI-generated summaries than human-expert-generated summaries. Therefore, AI summaries bridge the gap in comprehension while being particularly useful for those with less formal education, helping them grasp and engage with content that might otherwise be inaccessible.

\section{Discussion}\label{discussion}

Our pipeline shows how AI tools could help fill this gap. AI-generated summaries of judicial opinions score more favorably on readability metrics. In a survey experiment, we find that AI-generated summaries are clearer to participants. GPT-4's ability to summarize facts, define jargon, and contextualize legal arguments can help demystify court opinions. Style transfers into more accessible formats, e.g. for social media consumption, can ease the burden laypeople might face when reading court documents that are formatted with particular procedures and norms in mind.

That being said, there are a number of important caveats. First, AI is not a perfect substitute for lawyers. Simple prompts such as ``summarize this court opinion'' yield unsatisfactory results. Prompt engineering that draws on legal domain knowledge produces summaries that are clear and crisp. Hence, legal experts play a critical role in the development, evaluation, and deployment of AI assistants in legal applications. 

Second, there is an inherent tradeoff between fidelity to the original source material and simplification with any sort of summarization task. Our summaries are effective at conveying the main points of a legal argument but can miss some subtle nuances that legal professionals may find important. For example, a 7th-grade summary of \textit{Brown v. Board of Ed.} misses discussion of empirical evidence about the psychological impacts of segregation. It also misses the important historical context around how Chief Justice Earl Warren persuaded the rest of the Court to rule 9-0 in favor of Brown in part by convincing them that they did not need to explicitly overrule \textit{Plessy v. Ferguson} to reach the result. These limitations are to be expected as this additional context does not appear in the syllabus, but nonetheless demonstrates that a summary will not capture every legally relevant detail (Appendix B). 

Third, as with any machine learning approach, it is not realistic to expect ``perfect'' summaries. The ``ground truth'' of what constitutes a good summary will differ between contexts and even between two people. Even if there was an uncontroversial ground truth, a perfect accuracy score might indicate that the prompts are overfitted to the training data (the cases in our study) and might not generalize to all new cases.

There are several future research directions that emerge from this study. We are currently investigating whether simplified summaries that improve comprehension also serve a persuasive function - i.e., does reading a summary change one's mind about the outcome of a case? Developing methods for assessing the factual validity of legal concepts will be important for evaluating AI-assisted summaries at scale. Further work on legal AI prompt engineering and style transfers can open up different avenues for understanding how the public perceives legal summaries. A follow-up experiment might also test whether informing respondents about where a summary came from (e.g. from legal experts or AI) influences how they think about the summary. Finally, work on how disseminating easy-to-understand legal summaries and how this information affects public perceptions of courts and legal institutions will be of interest. 

\clearpage

\bibliography{bibliography}
\bibliographystyle{acl_natbib}

\clearpage
\newpage

\setcounter{table}{0}
\setcounter{figure}{0}
\setcounter{section}{0}
\renewcommand{\thetable}{A\arabic{table}}
\renewcommand{\thefigure}{A\arabic{figure}}
\renewcommand{\thesection}{\Alph{section}}

{\huge \noindent Appendix}

\section{Style-transfered Summaries}\label{app:style-transfer-summaries}

\begin{summaryquote}
    \textbf{7th Grade.} 1) In 2018, a law in Mississippi, known as the "Gestational Age Act," was challenged. This law banned most abortions after 15 weeks. The challenge came from the Jackson Women's Health Organization, the only licensed abortion clinic in the state. Lower courts blocked the law, saying there wasn't enough evidence that a fetus could survive outside the womb at 15 weeks. They also said that prior Supreme Court rulings didn't allow bans on abortions before a fetus could survive on its own.

2) The Supreme Court was asked to look at this case. They had to consider whether the Constitution, the highest law in the land, gives women a right to get an abortion. They decided that the Constitution does not. They disagreed with earlier decisions in the cases of Roe v. Wade and Planned Parenthood of Southeastern Pa. v. Casey, which had said that the Constitution did give women this right.

3) The Supreme Court also looked at whether the right to get an abortion is part of the "liberty" that the Constitution protects. The Constitution doesn't directly say that women have a right to get an abortion. But in the past, the Supreme Court had said that this right was part of a broader right to privacy. The new decision said that the Constitution doesn't protect this right.

4) The Supreme Court also looked at whether the right to get an abortion has deep roots in America's history and traditions. They decided that it does not. They pointed out that abortion was a crime in every state for a long time.

5) The Supreme Court also looked at whether the Constitution should protect a right just because the Supreme Court thinks Americans should have this freedom. They decided that the answer was no. The Supreme Court shouldn't decide what freedoms Americans should have based on their own opinions.

6) The Court also looked at whether the right to abortion was part of a larger right to make personal decisions without government interference. They concluded that this was not the case. They also argued that the right to an abortion is different from other rights because it involves the destruction of potential life.

7) The Supreme Court also considered whether they should keep following the decisions in Roe and Casey just because they were previous decisions. This is a legal principle called "stare decisis." They decided that they didn't have to keep following these decisions.

8) The Supreme Court also looked at whether the "undue burden" test from the Casey decision was a good way to decide whether a law about abortion was allowed. This test says that a law can't put a big obstacle in the way of a woman trying to get an abortion before the fetus can survive on its own. They decided that this test was not a good way to decide.

9) The Supreme Court also considered whether overruling Roe and Casey would cause problems because people have relied on these decisions. They decided that this wasn't a big concern. They said that getting an abortion isn't usually something people plan for, and that people could adjust quickly if states were allowed to ban abortions.

10) In the end, the Supreme Court decided that the Mississippi law was allowed. They said the Constitution doesn't give women a right to get an abortion, and that states can regulate or ban abortions. This was a big change from earlier decisions. They sent the case back to the lower courts to continue with this new understanding of the law.
\end{summaryquote}

\begin{summaryquote}
    \textbf{Twitter.} 1/10 This Supreme Court case concerns Mississippi's "Gestational Age Act," which bans most abortions post 15 weeks. It was challenged by Jackson Women's Health Organization and blocked by lower courts. The Supreme Court considers if the Constitution confers a right to abortion.

2/10 The Court suggests that Roe v. Wade* and Planned Parenthood v. Casey* did not fully answer if the Constitution confers a right to abortion. The Court concludes that it does not, overturning these precedents and returning abortion regulation to the people and their representatives.

3/10 The Court dissects whether the right to abortion is part of the "liberty" protected by the Fourteenth Amendment's Due Process Clause. It concludes that this right is not deeply rooted in the nation's history and tradition, thus it's not a substantive right protected by the Fourteenth Amendment.

4/10 The Court also rejects the argument that the Fourteenth Amendment's Equal Protection Clause supports the right to abortion. It concludes that a state's regulation of abortion is not a sex-based classification and is thus not subject to heightened scrutiny*.

5/10 The Court finds that the term "liberty" alone is not sufficient to protect a right to abortion. It asserts that history and tradition guide the interpretation of the Fourteenth Amendment, where there's no historical support for a constitutional right to abortion.

6/10 The Court stresses the historical context of the Fourteenth Amendment, noting that at the time of its adoption, most states had criminalized abortion. It argues that the consensus of state laws was misrepresented in Roe v. Wade and ignored in Casey v. Planned Parenthood.

7/10 The Court argues that the right to abortion is not part of broader entrenched rights like privacy or personal autonomy. It distinguishes abortion as unique because it involves potential life, which none of the other decisions cited by Roe and Casey involved.

8/10 The Court discusses the doctrine of stare decisis*, arguing that it doesn't counsel continued acceptance of Roe and Casey. While stare decisis is important for legal stability, it's not absolute and can be overruled in important constitutional decisions.

9/10 The Court identifies five factors that strongly favor overruling Roe and Casey: the nature of the Court's error, the quality of the reasoning in Roe, the unworkability* of the "undue burden" test, conflicts in the Courts of Appeals, and the distortion of related legal doctrines.

10/10 The Court concludes that the Constitution does not confer a right to abortion and that state regulations of abortion are subject to rational-basis review*. This case is returned to the lower courts for further proceedings consistent with this opinion.

*Glossary: \newline
Roe v. Wade and Casey refer to landmark abortion rights cases. \newline
Heightened scrutiny is a form of judicial review for laws potentially infringing on fundamental constitutional rights.\newline
Stare decisis requires courts to follow historical cases when making a ruling on a similar case. \newline
Workability refers to the practical applicability of a legal rule. \newline
Rational-basis review examines if a law is rationally related to a legitimate government interest
\end{summaryquote}

\begin{summaryquote}
    \textbf{YouTube.} 1) The Mississippi Gestational Age Act, which bans abortions after 15 weeks except for emergencies or severe fetal abnormalities, was challenged by Jackson Women's Health Organization. Lower courts sided with the clinic, citing lack of evidence for fetal viability and existing Supreme Court precedents against banning abortions prior to viability. However, the Supreme Court has overruled these decisions, arguing that the Constitution does not confer a right to abortion, and that authority to regulate abortion lies with the people and their elected representatives. 

2) The Court argues that the Constitution does not explicitly include a right to abortion. Although Roe v. Wade and Planned Parenthood v. Casey identified this right as part of a broader right to privacy, the Court rejects this interpretation, stating that the right to abortion is not deeply rooted in the nation's history or tradition. 

3) The Court examines the Fourteenth Amendment, which has sometimes been cited as a potential source of an implicit constitutional right to abortion. The Court rejects this argument, stating that abortion regulations are not a sex-based classification and are therefore not subject to heightened scrutiny.

4) The Court asserts that the Constitution does not protect the right to an abortion. It finds no basis for this right in American law, pointing out that abortion was a crime in every state until the latter part of the 20th century. 

5) The Court criticizes the Roe v. Wade decision for ignoring the historical context of the Fourteenth Amendment and misrepresenting the status of abortion as a common-law crime. It also rejects arguments that the right to abortion is part of a broader right to privacy or the freedom to make intimate and personal choices.

6) The Court argues that elected representatives should have the authority to regulate abortion, as the people of various states may have different views on the interests of a woman seeking an abortion and the interests of potential life. The Court also argues that the right to obtain an abortion cannot be justified as part of a broader right to autonomy, as this could potentially license fundamental rights to illicit activities.

7) The Court says that the doctrine of stare decisis, which emphasizes the importance of precedent in legal decision-making, does not necessitate continued acceptance of past rulings on abortion. Instead, precedent can be reevaluated in light of changing societal values and understandings.

8) The Court identifies five factors that favor overruling Roe and Casey, including the nature of the Court's error in earlier decisions, the quality of the reasoning in those decisions, and the effects of those decisions on related legal doctrines.

9) The Court criticizes the Roe v. Wade decision for conflating the right to privacy with the right to make personal decisions without governmental interference. It also argues that Roe v. Wade did not adequately consider the unique impact of abortion on potential life and failed to justify its distinction between pre- and post-viability abortions.

10) The Court concludes that the Mississippi Gestational Age Act is constitutional because it serves legitimate state interests and does not violate any fundamental constitutional right. The Court reverses the lower court's decision and remands the case for further proceedings.

*Rational-basis review: A standard of judicial review that examines whether a legislature had a reasonable and not arbitrary basis for enacting a particular statute.\newline
*Stare decisis: The legal principle of determining points in litigation according to precedent.
*Due Process Clause: A constitutional guarantee that no person shall be deprived of life, liberty, or property without due process of law.\newline
*Equal Protection Clause: A constitutional guarantee that no person or class of persons shall be denied the same protection of the laws that is enjoyed by other persons or other classes in like circumstances in their lives, liberty, property, and pursuit of happiness.\newline
*Heightened scrutiny: A form of judicial review that courts use to determine the constitutionality of certain laws.\newline
*Substantive rights: These are fundamental rights that the government cannot infringe upon, such as the right to free speech or the right to vote.\newline
*Common law: This is a body of unwritten laws based on legal precedents established by the courts.
*Scheme of ordered liberty: This phrase refers to the concept that individual liberty exists within a framework of established social order.\newline
*Quickening: The stage of pregnancy at which the mother can feel the movements of the fetus, traditionally considered to be around the middle of pregnancy.\newline
*Ordered liberty: A term used in constitutional law, referring to the balance between individual rights and social order.\newline
*Undue burden: A legal standard that prohibits laws imposing a substantial obstacle to a woman's choice to have an abortion.\newline
*Workability: The ability of a legal rule or standard to be understood and applied in a consistent and predictable manner.\newline
*Reliance interests: The expectations that parties have in the stability and predictability of the law, which can be disrupted when a court overrules a precedent.\newline
*Right to privacy: A legal concept that individuals have a protected right to privacy in their personal lives, even though this right is not explicitly mentioned in the Constitution.\newline
*Viability: In the context of abortion law, viability refers to the point at which a fetus is capable of living outside the womb.\newline
*Remand: To send a case back to a lower court for further action.
\end{summaryquote}

\section{7th-grade Level Summaries}

\noindent{\textbf{Bostock v. Clayton County (2020)}}
\begin{summaryquote}
1) Gerald Bostock, a man who is gay, was fired from his job in Clayton County, Georgia, after he joined a gay softball league. He had a good work record. He sued the county for firing him because he was gay. This is against Title VII of the Civil Rights Act of 1964. Lower courts dismissed his case, so he appealed to the Supreme Court.

2) This case is about three people who were fired from their jobs because of their sexual orientation or transgender identity. Gerald Bostock was fired for joining a gay softball league. Donald Zarda was fired after he told his employer he was gay. Aimee Stephens was fired after she announced she was going to transition from male to female.

3) The main issue is how to interpret Title VII of the Civil Rights Act of 1964. This law makes it illegal for employers to discriminate based on a person's sex, race, color, national origin, and religion.

4) This opinion argues that Title VII also makes it illegal for employers to fire someone because they are gay or transgender. When the law was passed, 'sex' meant the biological difference between males and females*. To 'discriminate' meant to treat someone unfairly. So, an employer who fires someone because of their sex is breaking the law. An employer can't avoid responsibility just by saying there were other reasons for the firing.

5) Cases involving gay or transgender discrimination are unique because there is usually more than one reason for the firing. This is why it's important to consider sex as a 'but-for' cause* in these cases. If an employer discriminates against a gay or transgender employee, sex is always a 'but-for' cause. This means the discrimination is intentional.

6) Three previous cases support this interpretation of Title VII. In Phillips v. Martin Marietta Corp, the court ruled that it doesn't matter how an employer tries to justify discrimination. What matters is the discriminatory action itself. In other words, if an employer treats an employee unfairly because of their sex, it's illegal, no matter what other reasons the employer gives.

7) The employers in this case want to use old societal norms to justify their actions. But the court should not be swayed by these arguments. What matters is whether the fired employee's sex was a 'but-for' cause of the firing. Claims that discrimination based on homosexuality or transgender status is different from sex discrimination are irrelevant. Discrimination against gay or transgender employees is always based on sex.

8) In conclusion, an employer who fires an employee because of their sex, regardless of any other factors, is breaking the law. The court should not accept any arguments that try to narrow the scope of Title VII. Such arguments go against the spirit of the law.

9) Some employers argue that because their policies apply to both men and women, they are not discriminatory. But this argument misses the point. The issue is not whether the policies apply to both sexes, but whether they result in unfair treatment based on sex.

10) In the end, we decided to reverse the lower courts' decisions and affirm the rights of the fired employees. Discrimination based on sex, including sexual orientation and transgender status, is illegal under Title VII.

'*Sex: In this context, 'sex' refers to the biological difference between males and females.

'*But-for cause: This is a legal term that means a specific action or condition was the direct cause of a certain outcome. In this case, it means that the employees' sex was the direct cause of their firing.

'*Discrimination: This means treating someone unfairly because of their sex, race, color, national origin, or religion.
\end{summaryquote}

\noindent{\textbf{Carpenter v. U.S. (2018)}}

\begin{summaryquote}
1) This case is about a man named Timothy Carpenter who was accused of robbery. The FBI used his cell phone records to show he was near the places where the robberies happened. Carpenter said the FBI should have gotten a warrant before getting his phone records. The lower court said it was okay for the FBI to get the records without a warrant, but the Supreme Court disagreed.

2) The Supreme Court said that getting Carpenter's cell phone records was like searching him, which the Fourth Amendment says can't be done without a warrant. The Fourth Amendment protects people's privacy and their stuff. The Court said that people should be able to expect that their privacy will be respected, and society agrees with this.

3) The Court said that the kind of data the FBI got from Carpenter's phone doesn't fit neatly into old rules. But, they said, tracking someone's past movements through their phone is a lot like tracking them with a GPS—it's detailed, it covers a lot of ground, and it's easy to do.

4) The Court didn't agree with the government's argument that the third-party doctrine* applies here. They said there's a big difference between the kind of information dealt with in old cases and the detailed location information that phone companies collect. They also said that this kind of information isn't really "shared" in the way we usually think of sharing.

5) The Court made it clear that this decision is narrow. It doesn't change the way old cases were decided or question the use of normal surveillance techniques. It also doesn't deal with other business records that might show where someone is, or with methods used for foreign affairs or national security.

6) In the end, the Court said that the government didn't get a warrant before getting Carpenter's phone records. They said that the law the government used to get the records, the Stored Communications Act, doesn't require the same level of proof as a warrant does.

*Cell-site location information (CSLI): This is a record that's made every time a cell phone connects to a cell tower. It shows where the phone is.

*Third-party doctrine: This is a legal idea that says if you give information to someone else, you can't expect it to stay private, even if you want it to.
\end{summaryquote}

\noindent{\textbf{Fisher v. University of Texas (2016)}}

\begin{summaryquote}
    1) Abigail Fisher, a white student, sued the University of Texas because she didn't get in. She said the school's policy of considering race in admissions was unfair and against the law. The Supreme Court disagreed with her.

2) The University of Texas has a two-part system for deciding who gets in. First, they automatically admit students who are in the top 10\% of their high school class. Second, they look at other factors for the rest of the students. One of these factors is race. They started this system in 2004 after a study showed that not considering race didn't give enough benefits of diversity to students.

3) Fisher wasn't in the top 10\% of her class, so she didn't get in. She said the school's policy of considering race was unfair to her and other white students. She said it went against the Equal Protection Clause*, which says everyone should be treated equally under the law. The case went all the way to the Supreme Court.

4) The Supreme Court said there are three important things to consider when looking at a school's policy on race. First, the school has to show that considering race is really important and legal. Second, while courts can review a school's policy, the school should have some freedom to decide what's best. Third, the school has to show that there are no other ways to achieve diversity without considering race.

5) The Supreme Court said Fisher didn't prove that the school treated her unfairly. She said the school should be clearer about how many minority students it wants to admit. But the Court said the goal isn't a specific number, but to make sure all students get the benefits of diversity.

6) The school has to have clear goals for diversity. The University of Texas met this requirement because it had clear, understandable goals based on data and legal positions.

7) Fisher said the school should use race-neutral methods, like it did before 2003. But the Court said data showed that these methods didn't work because there were still too few minority students. The school's policy of considering race helped make the freshman class more diverse and was necessary to prevent discrimination.

8) So, the Court said the school's policy was legal. It said considering race was necessary right now, but also gave guidelines for future legal challenges to affirmative action policies.

9) In conclusion, the Supreme Court ruled that the University of Texas's policy of considering race in admissions was legal. It said the policy was necessary to achieve diversity and prevent discrimination.

(*Equal Protection Clause: a part of the U.S. Constitution that says everyone should be treated equally under the law.)
\end{summaryquote}

\noindent{\textbf{Glacier Northwest Inc. v. International Brotherhood of Teamsters (2023)}}

\begin{summaryquote}
    1) This case is about a company called Glacier Northwest and a union called Local 174. The company sued the union because of damage done during a strike. The case looked at whether the union's actions were protected by a law called the National Labor Relations Act (NLRA)*.

2) Glacier Northwest said the union damaged their property on purpose during the strike. The union's truck drivers stopped working and didn't take steps to stop concrete from hardening and becoming useless. This is not what usually happens during a strike.

3) The company said the union's actions were planned and harmful. They accused the union of conversion* and trespass to chattels*. But the state court said the union's actions were protected by their right to organize and strike.

4) But Glacier Northwest argued that just because the union has the right to strike, it doesn't mean they can damage the company's property. This is not considered ethical in labor law.

5) The union's actions seemed planned. They went on strike at a time when it would cause a lot of damage to the company's materials and equipment. This raises questions about whether these actions are allowed under the NLRA.

6) The court looked at whether the union's actions were planned and whether they gave the company enough warning before the strike. These factors didn't take away the union's rights under federal law, but they did play a role in deciding who was responsible for the damage.

7) The union not only risked damaging the company's trucks with the hardened concrete, but they also seemed to trick the company before the strike. If these allegations are true, this goes against the spirit of the NLRA.

8) The court decided that the union's actions were not protected by the NLRA. This means that the company's claims were not preempted* by the NLRA, as other courts had said.

9) The court's decision shows that while unions have the right to strike, they can't damage a company's property on purpose. This is not part of their rights under the NLRA.

10) In conclusion, the Supreme Court ruled in favor of Glacier Northwest. They said the union's actions were not protected by the NLRA and that the company's claims were valid. This means that the union could be held responsible for the damage done during the strike.

*National Labor Relations Act (NLRA): A law that protects the rights of employees and employers, encourages collective bargaining, and limits certain labor and management practices that can harm the general welfare of workers, businesses, and the U.S. economy.
*Conversion: When someone interferes with another person's personal property.
*Trespass to chattels: When someone unlawfully interferes with someone else's personal property.
*Preempted: When a higher law (like a federal law) takes precedence over a lower law (like a state law).
\end{summaryquote}

\noindent{\textbf{Gonzalez v. Carhart (2007)}}

\begin{summaryquote}
    1) This case, Gonzales v. Carhart, is about a law called the "Partial-Birth Abortion Ban Act of 2003." This law was made to stop a type of abortion where the baby dies when part of it is outside the mother's body. Dr. Leroy Carhart and other doctors said this law was not fair because it could stop other types of abortions too. They said this law was against a woman's right to choose to have an abortion.

2) The law only stops a certain type of abortion that happens late in pregnancy. It does not stop the most common types of abortions that happen early in pregnancy. There are two main ways to do an abortion late in pregnancy. One way is called 'standard dilation and evacuation' (D\&E), where the baby is taken apart and removed piece by piece. The other way is called 'intact D\&E,' where the baby is removed all in one piece. This second way is more complicated and can be more dangerous for the woman.

3) The government made this law because they believe in the value of human life. They also want to make sure doctors are doing what is best for their patients. The law is very clear about what it means by 'partial-birth abortion.' It also makes sure that doctors can still do other types of abortions if the woman's health is at risk.

4) Both sides in the case had good points, but neither side was able to convince the court completely. The government said they had the right to make this law, and the doctors said the law could put women's health at risk. The court said that a woman could still challenge the law if she needed to have the type of abortion that the law bans.

5) In the end, the court decided that the law was okay. They said the law was a good way to deal with the issue of abortion. They understood that people have different opinions about abortion, but they thought the law was a fair way to handle these differences.

* Penumbras: This is a fancy word for the shadows or partially colored areas you see during an eclipse.
* Incorporation: This is the idea that some parts of the U.S. Constitution also apply to state laws.
* Miranda rights: These are the rights you have when you are arrested. They include the right to stay quiet and the right to have a lawyer.
* Strict scrutiny: This is a way for courts to look at laws that might be unfair to certain groups of people. The government has to have a very good reason for making these laws, and the laws have to be the best way to solve the problem.
\end{summaryquote}

\noindent{\textbf{Harris v. Quinn (2014)}}

\begin{summaryquote}
    1) This case is about a group of personal care assistants (PAs) in Illinois who sued the governor and some unions. They said a law that made them pay fees to the union, even if they didn't want to join, was against their rights. The lower courts didn't agree with them, but the Supreme Court did.

2) The law they were fighting against came from an older case, Abood v. Detroit Bd. of Ed. This case said that unions could charge fees to everyone, even if they didn't want to join. The fees were supposed to help with things like negotiating better pay and benefits.

3) But the Supreme Court said this old case didn't really apply to the PAs. They said PAs were different from regular state employees because they didn't rely on the state as much for their jobs. Instead, they mostly answered to the people they were helping.

4) The Court also said that the union couldn't do as much for the PAs as it could for regular state employees. So, it didn't make sense to make the PAs pay the same fees.

5) The people defending the law said it helped keep peace and was good for the PAs. But the Court didn't agree. They said the PAs didn't work together in a state facility and the union couldn't do much for them. So, the law didn't really help keep peace.

6) In the end, the Court decided that the fees were a bigger problem than any benefits they might bring. They said the fees went against the PAs' right to choose who they associate with.

7) The Court also said the other arguments for the law didn't make sense in this case. They said the facts were different, so the old cases didn't apply.

8) So, the Court ruled that the law was against the First Amendment, which protects people's freedom to choose who they associate with. They said the state couldn't take money from the PAs' paychecks to support the union if the PAs didn't want to join.

9) This decision was made using a rule called *strict scrutiny. This rule says the government has to have a really good reason to limit people's rights and it has to do it in the best way possible.

10) In this case, the Court said the government didn't have a good enough reason to take money from the PAs' paychecks. And even if it did, taking the money wasn't the best way to do it.

*Strict scrutiny: This is a rule the courts use when they're deciding if a law goes against people's rights. The government has to show that the law is really important and that it's the best way to achieve its goal.
\end{summaryquote}

\noindent{\textbf{Janus v. AFSCME (2018)}}

\begin{summaryquote}
    1) This case is about a man named Janus who works for the government in Illinois. He doesn't want to be in a union, but the law says he has to pay fees to the union anyway. The Supreme Court had to decide if this law is fair.

2) In Illinois, if most workers vote to have a union, then all workers have to follow the union's rules. This includes paying a fee every year. The union says what the fee is for in a notice.

3) Janus thinks this isn't fair. He says it's against his First Amendment rights, which protect freedom of speech. He doesn't want to pay for the union's bargaining, especially if he disagrees with what they're bargaining for.

4) The Supreme Court had to think about a past case, Abood v. Detroit Bd. of Ed, which said these fees were okay. But most of the justices didn't think that case applied here.

5) The old case said the fees were okay for two reasons: they keep peace at work and they stop "free riders."* Free riders are people who get benefits without paying for them. But the justices found evidence that peace at work can happen without these fees. They also didn't think stopping free riders was a good enough reason to limit free speech.

6) The justices decided that the fees don't pass the tests for limiting First Amendment rights. These tests are called 'Standard,' 'Strict,' and 'Exacting Scrutiny.'** They also thought it was hard to tell what the union was really spending the fees on. And they thought the old case was too lenient about First Amendment rights.

7) The justices said that workers have to agree to pay the fees. They can't be forced to pay. This decision overruled the old case and the Illinois law. The justices said forcing workers to pay the fees was against their First Amendment rights.

8) So, in the end, the Supreme Court sided with Janus. They said that forcing workers to pay union fees, even if they don't want to be in the union, is against their First Amendment rights.

*Free riders: People who get benefits from something without paying for it.

**Standard/Strict/Exacting Scrutiny: These are tests to see if a law is fair. 'Strict scrutiny' is the toughest test. It means the law has to be the least restrictive way to achieve a goal. 'Exacting scrutiny' looks closely at whether the law balances free speech rights with other important goals. 'Standard scrutiny' means the government can regulate something if it can show a good reason for it.
\end{summaryquote}

\noindent{\textbf{Obergefell v. Hodges (2015)}}

\begin{summaryquote}
    1) This case is about same-sex marriage. Same-sex couples in Ohio, Michigan, Kentucky, and Tennessee sued their states. They said the states were wrong to ban same-sex marriage or not recognize legal same-sex marriages from other states. They said this violated the Equal Protection and Due Process Clauses of the Fourteenth Amendment*. At first, they won. But then the U.S. Court of Appeals for the Sixth Circuit said the bans did not violate the couples' Fourteenth Amendment rights.

2) The couples said the laws in Michigan, Kentucky, Ohio, and Tennessee were unfair. These laws defined marriage as between one man and one woman. The couples said these laws violated the Fourteenth Amendment. They said they had the right to marry or have their out-of-state marriages recognized in the states where they lived.

3) At first, the courts agreed with the couples. They said the couples had a fundamental right to marry and have their out-of-state marriage recognized. But then the Sixth Circuit overruled these decisions. The question now is whether the Fourteenth Amendment requires a State to license a same-sex marriage and recognize a lawful, out-of-state same-sex marriage.

4) The couples argued that the laws did not respect their personal dignity and autonomy. They said past court decisions have protected personal choices that are important to a person's identity. They said the Fourteenth Amendment’s Due Process Clause protects these choices. They said the State has a responsibility to respect these choices.

5) The couples said there are four important principles about marriage. They said people have the right to choose who they marry. They said same-sex couples should have the same legal rights as opposite-sex couples. They said protecting children and families is important. They said marriage is an important part of our society.

6) Sometimes, people's understanding of things can change. They can see that something they thought was okay is actually unfair. This can happen when people learn more about a social issue. The couples said this is what happened with same-sex marriage. They said people now understand that not allowing same-sex marriage is unfair.

7) Some people think marriage should only be between a man and a woman. But the couples said this view is biased. They said it does not respect the fundamental right to marriage.

8) Freedom and equality are important principles in our Constitution. They may not mean the same thing in every situation. But they are important in this case. They are important when we think about how we should treat homosexual individuals. They are also important when we think about how we should protect religious beliefs.

9) The majority of people believe in the importance of marriage. But the laws in these states do not respect this belief. They do not allow same-sex couples to marry. They do not recognize same-sex marriages from other states. This is unfair.

10) The Supreme Court ruled that the laws in these states are wrong. They said the Fourteenth Amendment requires a State to license a same-sex marriage and recognize a lawful, out-of-state same-sex marriage. They said these laws do not respect the personal dignity and autonomy of same-sex couples. They said these laws do not protect the fundamental right to marriage.

*The Fourteenth Amendment is a part of the U.S. Constitution. It says that all people have equal protection under the law. The Equal Protection and Due Process Clauses are parts of the Fourteenth Amendment. They say that all people have the right to fair treatment and due process of law.

\end{summaryquote}

\noindent{\textbf{Riley v. California (2014)}}

\begin{summaryquote}
    1) This case is about a man named David Riley who was arrested because the police found guns in his car. The police also looked through his cell phone without a warrant and found evidence that he was part of a gang. This evidence was used to connect him to a shooting and he was sentenced to 15 to life in prison. The Supreme Court had to decide if the police were allowed to search his phone without a warrant.

2) The Supreme Court had to look at two cases, RILEY v. CALIFORNIA and United States v. Wurie. Both cases were about whether the police could look through the information on a cell phone without a warrant after arresting someone. In both cases, the evidence found on the phones led to the defendants being convicted.

3) The Fourth Amendment of the Constitution says that people are protected from unreasonable searches and seizures. Usually, the police are allowed to search someone without a warrant if they have been arrested. This is to keep the police safe and to make sure evidence isn't destroyed. But this only applies to the person being arrested and the area around them.

4) The Court decided that the rules for physical searches don't apply to cell phones. They pointed out that cell phones are different because they can store a lot of personal information. This raises concerns about privacy.

5) The Court said that the reasons for allowing warrantless searches don't apply to cell phones. Digital data on a phone doesn't pose a threat to police safety and it's not likely to be destroyed quickly. If there is an immediate danger, the police can use *exigent circumstances to search the phone.

6) The Court also said that looking through someone's digital data is different from searching their pockets. It's a bigger invasion of privacy. Plus, data on a phone can be stored on a remote server, which means a search could reach beyond the immediate area.

7) The government argued that if there's probable cause, a warrantless search is okay. But the Court disagreed. They said that the invasion of privacy from a casual arrest is too big compared to the smaller interests of the Fourth Amendment.

8) The government also suggested limiting the scope of the search. But the Court said this was flawed. It would violate the *categorical-rule law and force the court to decide what's okay and what's not in a digital search. This could lead to mistakes.

9) The Court decided that the police can still get the information on a cell phone, but they need to get a warrant first. They said that warrants are easy to get these days. There might be some rare cases (**exigent circumstances) where the police can search a phone without a warrant.

10) In conclusion, the Supreme Court decided that the police need a warrant to search a cell phone, even if the person has been arrested. This protects people's privacy and follows the Fourth Amendment.

*Exigent circumstances: This is when the police can act quickly in an emergency without a warrant.
*Categorical-rule law: This is when the court uses strict rules to guide decisions.
**Exigent circumstances: This is when there's an emergency and the police need to act quickly without a warrant.
\end{summaryquote}

\noindent{\textbf{Schuette v. Coalition to Defend Affirmative Action}}

\begin{summaryquote}
    1) The case of Schuette v. Coalition to Defend Affirmative Action is about whether voters in Michigan can decide to stop using race as a factor in college admissions. This is called affirmative action. Some people said this was unfair and against the Equal Protection Clause*, which says everyone should be treated equally.

2) Before this case, courts had said that the University of Michigan couldn't use race to decide who gets into their school. So, voters in Michigan passed Proposal 2. This law said that public colleges in Michigan couldn't use race as a factor in admissions. 

3) Some groups didn't like this law. They said it was against the idea of equal opportunity. They also said it was against the 14th Amendment*, which says that laws should be fair to everyone, no matter their race.

4) The main question in this case was about the 14th Amendment. This amendment says that laws can't treat people differently because of their race. But it also says that laws can't ignore race completely. This is called 'strict scrutiny'*.

5) The people who didn't like Proposal 2 said it was like other laws that had been unfair to certain races. But the court said Proposal 2 was different. It didn't say that one race was better than another. It just said that race couldn't be a factor in college admissions.

6) The court also said that it was important not to group people by race. They said this could lead to decisions that were unfair to certain races. This is called 'questionable constitutionality'*.

7) The court said that Proposal 2 didn't go against the Equal Protection Clause. They said it didn't treat any race unfairly. It just said that race couldn't be a factor in college admissions.

8) The court also said that laws can't ignore race completely. But they also can't treat people differently because of their race. This is a difficult balance to find.

9) In the end, the court said that it was up to the voters to decide. They said that the voters had the right to decide whether race should be a factor in college admissions.

10) So, the court decided that Proposal 2 was okay. They said it didn't go against the Equal Protection Clause or the 14th Amendment. They said it was up to the voters to decide this issue.

*Equal Protection Clause: This is a part of the 14th Amendment that says everyone should be treated equally by the law.
*14th Amendment: This is a part of the Constitution that says everyone should be treated equally by the law, no matter their race.
*Strict scrutiny: This is a way for courts to look at laws that treat people differently because of their race. The law has to have a very good reason for treating people differently.
*Questionable constitutionality: This is when a law might go against the Constitution.
\end{summaryquote}

\noindent{\textbf{Students for Fair Admissions v. Harvard (2023)}}

\begin{summaryquote}

1) This case is about a group called Students for Fair Admissions (SFFA) who sued Harvard College and the University of North Carolina (UNC). They said that these schools were not fair in their admissions process because they were using race as a factor, which they believed was against the law. The law they referred to is the Civil Rights Act of 1964 and the Equal Protection Clause of the Fourteenth Amendment*.

2) The Equal Protection Clause is a part of the Fourteenth Amendment that says that every person should be treated equally by the law, no matter their race, color, or nationality. The SFFA believed that by considering race in admissions, Harvard and UNC were not treating all applicants equally.

3) The Court looked at the history of the Fourteenth Amendment and how it has been used in the past. They also looked at how other cases involving race and college admissions were handled. They found that while diversity in a student body can be a good thing, it must be handled in a way that treats all applicants fairly and equally.

4) The Court also looked at the idea of "strict scrutiny*". This is a way for the courts to look at laws to see if they are fair and necessary. If a law or policy is found to be unfair or unnecessary, it may not pass strict scrutiny and could be considered unconstitutional.

5) The Court found that the admissions systems at Harvard and UNC did not pass strict scrutiny. They said that the schools' use of race in admissions was not clear or specific enough, and it resulted in fewer admissions for certain racial groups. They also said that the schools' use of race in admissions seemed to stereotype certain racial groups, which is not allowed.

6) The Court also said that the schools' admissions systems did not have a clear end point. This means that there was no clear plan for when the schools would stop using race as a factor in admissions. This was another reason why the Court said the schools' admissions systems were not fair.

7) The Court decided that the admissions systems at Harvard and UNC were not fair and did not follow the Equal Protection Clause of the Fourteenth Amendment. They said that the schools' use of race in admissions was not clear, specific, or fair enough to be allowed.

8) However, the Court also said that schools can consider how race has affected an applicant's life. They can look at how an applicant's experiences with their race have shaped them and what they can bring to the school because of those experiences.

9) In the end, the Court decided that the admissions systems at Harvard and UNC were not fair and did not follow the law. They said that the schools' use of race in admissions was not allowed because it was not clear, specific, or fair enough.

10) So, the Court decided that the SFFA was right. They said that Harvard and UNC were not treating all applicants equally in their admissions process, which is against the law. They said that the schools needed to change their admissions systems to be fair to all applicants, no matter their race.

*The Equal Protection Clause is a part of the Fourteenth Amendment that says that every person should be treated equally by the law, no matter their race, color, or nationality.
*Strict scrutiny is a way for the courts to look at laws to see if they are fair and necessary. If a law or policy is found to be unfair or unnecessary, it may not pass strict scrutiny and could be considered unconstitutional.
\end{summaryquote}

\noindent{\textbf{U.S. v. Windsor (2013)}}

\begin{summaryquote}
    1) The Supreme Court case United States v. Windsor is about a woman named Edith Windsor. She had to pay a lot of money in taxes when her wife died because the federal government didn't recognize their marriage. She said this was unfair and against the Constitution.

2) The Defense of Marriage Act (DOMA)* was a law that said marriage was only between a man and a woman. Because of this law, Edith couldn't get a tax break that other married people could get when their spouse died. She said this was against the Fifth Amendment, which says everyone should be treated equally.

3) This case was complicated because many different groups were involved. The Supreme Court had to decide if it had the power to make a decision about this case. They said they did because they needed to solve a real problem between two sides.

4) Before this case, the Supreme Court usually let states decide what marriage was. But in this case, they said DOMA was wrong because it hurt people that the state of New York was trying to protect. They said this was against the Fifth Amendment.

5) DOMA made things confusing because it said marriage was one thing, but some states said it was something else. This made things hard for people who were married in their state but not according to federal law. The Supreme Court said this wasn't fair and went against the tradition of states deciding what marriage is.

Terms:

1) Defense of Marriage Act (DOMA): A law that said marriage was only between a man and a woman.
2) Bipartisan Legal Advisory Group (BLAG): A group of five people in the House of Representatives who decide on legal issues that affect the House.
3) Article III jurisdiction requirement: The rule that courts can only decide on real problems between two sides.
4) Prudential limitations: Rules that courts make up to decide on certain types of problems.
5) Loving v. Virginia, Sosna v. Iowa, Ohio ex rel.Popovici v. Agler: Past court cases that said states can decide what marriage is, as long as they treat everyone equally.
\end{summaryquote}

\noindent{\textbf{U.S. v. Jones (2012)}}

\begin{summaryquote}
    1) This case is about Antoine Jones, who was arrested for having drugs. The police found the drugs by tracking his car with a GPS device, but they didn't have a proper warrant to do this. The case is about whether the police broke the Fourth Amendment* rule against unreasonable searches and seizures.

2) The police had a warrant to put the GPS device on the car, but they didn't follow the rules of the warrant. They put the device on the car a day late and in a different place than the warrant allowed.

3) The GPS device was on the car for four weeks. It helped the police find evidence to charge Jones with drug trafficking. The court said the police couldn't use the GPS data from when the car was parked at Jones's house because that was his private space.

4) But the court said the police could use the GPS data from when the car was on public roads. The court said Jones didn't have a 'reasonable expectation of privacy*' when he was driving on public roads. The D.C. Circuit court disagreed and said using the GPS device without a proper warrant broke the Fourth Amendment rule.

5) The main question is whether using a GPS device to track a car without a proper warrant is an 'unreasonable search and seizure*' under the Fourth Amendment. It's clear that it is. The Fourth Amendment was made to protect people's private spaces from being disturbed without good reason.

6) Some people might say that the Fourth Amendment only protects against physical trespassing. But in the mid-20th century, the Supreme Court started to see the Fourth Amendment as protecting people's reasonable expectations of privacy, not just their physical property.

7) The important thing is that the Fourth Amendment protects people from the government messing with their stuff. This has been true since the Fourth Amendment was first made. The idea of 'reasonable expectation of privacy' doesn't change this.

8) Some people might say that even if tracking the car was a 'search,' it was a 'reasonable' one. But this argument wasn't made in the lower courts, so it doesn't matter here.

9) In conclusion, the police broke the Fourth Amendment rule when they tracked Jones's car without a proper warrant. They can't do this without a good reason and proper permission.

*Definitions:
1. Fourth Amendment: A rule in the U.S. Constitution that protects people from unreasonable searches and seizures by the government.
2. 'Reasonable expectation of privacy': The idea that people should be able to expect a certain level of privacy in certain situations.
3. 'Unreasonable searches and seizures': When the government searches or takes someone's property without a good reason.
4. Common-Law trespass: When someone goes onto someone else's property without permission.
\end{summaryquote}

\noindent{\textbf{Whole Woman's Health v. Hellerstedt (2016)}}

\begin{summaryquote}
    1) This case is about a law in Texas that made it harder for abortion clinics to operate. The law said that doctors who perform abortions must be able to admit patients to a nearby hospital, and that clinics must meet the same standards as surgery centers. Many clinics had to close because of these rules. A group of abortion providers sued, saying the law was against the Constitution.

2) The Supreme Court had to decide if the Texas law was an "undue burden" on a woman's right to have an abortion. An "undue burden" means something is too hard or unfair. The Court had to look at how the law affected women, like if they had to travel far to get to a clinic or wait a long time for an appointment.

3) The Court said the lower court made a mistake when it treated this case like it was about economic laws, not constitutional rights. The Court also said Texas didn't prove that the law actually made women's health better.

4) The Court found that the law did create an undue burden. Even though each rule might not seem too hard on its own, all of them together made it too difficult for women to get an abortion. The law also made it very expensive for clinics to stay open.

5) The Court also said that the abortion providers could bring up new arguments in this case, even though they had already sued Texas before. This is because the situation had changed since the law was put into effect.

6) The Court pointed out that the lower court didn't understand some important things about the case. For example, it didn't realize how much the law was hurting women's rights.

7) The Court's decision was about finding a balance. It wanted to protect women's rights and make sure they could get medical care. It also wanted to make sure the law was fair and didn't punish women for not having access to a clinic.

*Undue burden: This is a legal term that means a law or rule is too hard or unfair. If a law creates an undue burden, it might be unconstitutional.
*Facial challenge: This is when someone says a law is always unconstitutional, no matter how it's used. In this case, the abortion providers said the Texas law was always unconstitutional because it always made it too hard for women to get an abortion.
\end{summaryquote}
\section{Brown v. Board of Ed. Summary}\label{app:brown-v-board-summary}

\begin{summaryquote}
1) This case is about a group of African American students who were not allowed to go to certain public schools because of laws that said it was okay to separate people by race. The students said this was not fair and went against the Fourteenth Amendment*, which says everyone should be treated equally. But, the lower courts said no because of an old rule that said it was okay to separate people by race as long as the schools for each race were equal.

2) The main point of the case is about whether it's okay to separate white and black children in public schools just because of their race. The majority of the judges say that this is not okay, even if the schools for white and black children have the same things like books, desks, and teachers.

3) The judges first talk about the history of the Fourteenth Amendment. They say it's not clear what the people who made the amendment thought about public schools. But, they think that we should not just look at what people thought back then. We should also think about how important public schools are now.

4) The judges also say that when a state decides to have public schools, it has to let everyone go to those schools on the same terms. They believe that separating children in public schools just because of their race is not fair. It takes away from the black children's chance to have the same educational opportunities as white children, even if the schools have the same things.

5) The judges strongly disagree with the old rule from Plessy v. Ferguson that said it was okay to separate people by race as long as the separate places were equal. They say this rule does not apply to public schools. They believe that this rule is not fair and goes against the Fourteenth Amendment's promise to treat everyone equally.

6) In the end, the judges decide to look at the case again to answer more questions about how to make things right.

*Fourteenth Amendment: This is a change to the U.S. Constitution that says everyone should be treated equally under the law. This means that states can't make laws that treat some people differently than others.
\end{summaryquote}

\clearpage

\begin{table}[]
    \centering
        \caption{List of Selected Cases in our Survey Experiment}
    
    \begin{tabular}{l l}
    Area of Law & Case \\ \hline
    \multirow{3}{*}{Abortion Rights} & Gonzalez v. Carhart (2007) \\
    & Whole Woman's Health v. Hellerstadt (2016) \\
    & Dobbs v. Jackson (2022) \\ \hline
    \multirow{3}{*}{Affirmative Action} & Schuette v. Coalition to Defend Affirmative Action (2014) \\
    & Fisher v. University of Texas (2016) \\
    & Students for Fair Admission v. Harvard (2023) \\ \hline
    \multirow{3}{*}{Labor} & Harris v. Quinn (2014) \\
    & Janus v. AFSCME (2018) \\
    & Glacier Northwest v. Teamsters (2023) \\ \hline
    \multirow{3}{*}{LGBT Rights} & U.S. v. Windsor (2013) \\
    & Obergefell v. Hodges (2015) \\
    & Bostock v. Clayton Country (2020) \\ \hline
    \multirow{3}{*}{Search \& Seizure} & U.S. v. Jones (2012) \\
    & Riley v. California (2014) \\
    & Carpenter v. U.S. (2018) \\   \hline
    \end{tabular}

    \label{tab:selected-cases}
\end{table}

\begin{table}[]
    \centering
        \caption{Regression Results}
\input{figures/main.tex}
\flushleft {\footnotesize Notes. Main results for estimated treatment effect of our AI summaries compared to control (Justia summaries). Coefficients are presented with standard errors in parentheses. Significance levels are indicated as follows: * p < 0.05, ** p < 0.01, *** p < 0.001.} 
\label{tab:main}
\end{table}

\end{document}

%% file: figures/main.tex
{
\def\sym#1{\ifmmode^{#1}\else\(^{#1}\)\fi}
\small
\begin{tabular}{l*{6}{c}}
\hline\hline
                         &\multicolumn{1}{c}{(1)}&\multicolumn{1}{c}{(2)}&\multicolumn{1}{c}{(3)}&\multicolumn{1}{c}{(4)}&\multicolumn{1}{c}{(5)}&\multicolumn{1}{c}{(6)}\\
                         &\multicolumn{1}{c}{Heard of Case}&\multicolumn{1}{c}{Case Area Correct}&\multicolumn{1}{c}{Case Decision Correct}&\multicolumn{1}{c}{Level of Detail}&\multicolumn{1}{c}{Clarity}&\multicolumn{1}{c}{Share with Friend}\\
\hline
AI Summary               &    0.0385         &   0.00192         &     0.107\sym{**} &     0.202\sym{***}&     0.431\sym{***}&     0.432\sym{***}\\
                         &  (0.0308)         &  (0.0157)         &  (0.0354)         &  (0.0384)         &  (0.0752)         &  (0.0911)         \\
\hline
Observations             &       560         &       560         &       560         &       560         &       560         &       560         \\
\hline\hline
\end{tabular}
}